\documentclass{icga}
\usepackage[dvips]{graphicx}

\newcommand{\tab}{\ \ \ \ \ \ }

\title{VERIFIED NULL-MOVE PRUNING}
\runningtitle{Verified Null-Move Pruning}

\author{Omid David-Tabibi \thanks{Department of Computer Science,
Bar-Ilan University, Ramat-Gan 52900, Israel, Email:
mail@omiddavid.com, Web: http://www.omiddavid.com.} \and
Nathan S. Netanyahu
\thanks{Department of Computer Science, Bar-Ilan University,
Ramat-Gan 52900, Israel, Email: nathan@cs.biu.ac.il, and Center
for Automation Research, University of Maryland, College Park, MD
20742, USA, Email: nathan@cfar.umd.edu. }}

\affiliation{Ramat-Gan, Israel}
\issue{September 2002}

\begin{document}


\maketitle 

\begin{abstract}
In this article we review standard null-move pruning and introduce
our extended version of it, which we call \emph{verified null-move
pruning}. In verified null-move pruning, whenever the shallow null-move
search indicates a fail-high, instead of cutting off the search from the
current node, the search is continued with reduced depth.

Our experiments with verified null-move pruning show that on average, it
constructs a smaller search tree with greater tactical strength in
comparison to standard null-move pruning. Moreover, unlike standard
null-move pruning, which fails badly in zugzwang positions, verified
null-move pruning manages to detect most zugzwangs and in such
cases conducts a re-search to obtain the correct result. In addition,
verified null-move pruning is very easy to implement, and any standard
null-move pruning program can use verified null-move pruning by
modifying only a few lines of code.
\end{abstract} \vspace*{-10pt}

\section{Introduction}

Until the mid-1970s most chess programs were trying to search the
same way humans think, by generating ``plausible'' moves. By using
extensive chess knowledge at each node, these programs selected a
few moves which they considered plausible, and thus pruned large
parts of the search tree. However, plausible-move generating
programs had serious tactical shortcomings, and as soon as
brute-force search programs like \textsc{Tech}
\citebay{gillogly72} and \textsc{Chess 4.x} \citebay{slate77}
managed to reach depths of 5 plies and more, plausible-move
generating programs frequently lost to brute-force searchers due
to their tactical weaknesses. Brute-force searchers rapidly
dominated the computer-chess field.

Most brute-force searchers of that time used no selectivity in
their full-width search tree, except for some extensions,
consisting mostly of check extensions and recaptures. The most
successful of these brute-force programs were \textsc{Belle}
(Condon and Thompson, 1983a,b)\nocite{condon83_1}, \textsc{Deep
Thought} \citebay{hsu90}, \textsc{Hitech}
\citebay{berliner90,berliner87,ebeling86}, and \textsc{Cray Blitz}
\citebay{hyatt90}, which for the first time managed to compete
successfully against humans.

The introduction of null-move pruning
\citebay{beal89,goetsch90,donninger93} in the early 1990s marked
the end of an era, as far as the domination of brute-force
programs in computer chess is concerned. Unlike other
forward-pruning methods (e.g., \emph{razoring}
\citebay{birmingham77}, \textsc{Gamma} \citebay{newborn75}, and
\emph{marginal forward pruning} \citebay{slagle71}), which had
great tactical weaknesses, null-move pruning enabled programs to
search more deeply with minor tactical risks. Forward-pruning
programs frequently outsearched brute-force searchers, and started
their own reign which has continued ever since; they have won all
World Computer-Chess Championships since 1992
\citebay{herik92,tsang95,feist99}. \textsc{Deep Blue}
\citebay{hammilton97,hsu99} (the direct descendant of \textsc{Deep
Thought} \citebay{hsu90}) was probably the last brute-force
searcher. Today almost all top-tournament playing programs use
forward-pruning methods, null-move pruning being the most popular
of them \citebay{feist99}.

In this article we introduce our new \emph{verified null-move
pruning} method, and demonstrate empirically its improved
performance in comparison with standard null-move pruning. This is
reflected in its reduced search tree size, as well as its greater
tactical strength. In Section 2 we review standard null-move
pruning, and in Section 3 we introduce verified null-move pruning.
Section 4 presents our experimental results, and Section 5
contains concluding remarks.

\section{Standard Null-Move Pruning}

As mentioned earlier, brute-force programs refrained from pruning any
nodes in the full-width part of the search tree, deeming the risks of
doing so as being too high. Null-move \citebay{beal89,goetsch90,donninger93}
introduced a new pruning scheme which based its cutoff decisions on
dynamic criteria, and thus gained greater tactical strength in comparison
with the static forward pruning methods that were in use at that time.

Null-move pruning is based on the following assumption: in every
chess position, doing nothing (i.e., doing a null move) would not
be the best choice even if it were a legal option. In other words,
the best move in any position is better than the null move. This
idea enables us easily to obtain a lower bound $\alpha$ on the
position by conducting a null-move search. We make a null move,
i.e., we merely swap the side whose turn it is to move. (Note that
this cannot be done in positions where that side is in check,
since the resulting position would be illegal. Also, two null
moves in a row are forbidden, since they result in nothing.) We
then conduct a regular search with reduced depth and save the
returned value. This value can be treated as a lower bound on the
position, since the value of the best (legal) move has to be
better than that obtained from the null move. If this value is
greater than or equal to the current upper bound (i.e., $value \ge
\beta$), it results in a cutoff (or what is called a fail-high).
Otherwise, if it is greater than the current lower bound $\alpha$,
we define a narrower search window, as the returned value becomes
the new lower bound. If the value is smaller than the current
lower bound, it does not contribute to the search in any way. The
main benefit of null-move pruning is due to the cutoffs, which
result from the returned value of null-move search being greater
than the current upper bound. Thus, the best way to apply
null-move pruning is by conducting a minimal-window null-move
search around the current upper bound $\beta$, since such a search
will require a reduced search effort to determine a cutoff. A
typical null-move pruning implementation is given by the
pseudo-code of Figure~\ref{fig:std-algo}.

\begin{figure}[h!]
\begin{center}
\begin{tabular}{|l|}
\hline
{\sl /* the depth reduction factor */}\\ {\tt \#define R \ 2}\\
{\tt int search (alpha, beta, depth)} \{\\ \tab {\tt if (depth
$<$= 0)}\\ \tab \tab {\tt return evaluate();} {\sl /* in practice,
quiescence() is called here */} \\ \tab {\sl /* conduct a
null-move search if it is legal and desired */}\\ \tab {\tt if
($!$in\_check() \&\& null\_ok())}\{\\ \tab \tab {\tt
make\_null\_move();}\\ \tab \tab {\sl /* null-move search with
minimal window around beta */}\\ \tab \tab {\tt value =
-search(-beta, -beta + 1, depth - R - 1);}\\ \tab \tab {\tt if
(value $>$= beta)} {\sl /* cutoff in case of fail-high */}\\ \tab
\tab \tab {\tt return value;}\\ \tab \}\\ \tab {\sl /* continue
regular NegaScout/PVS search */}\\ \tab \ldots \\ \}\\
\hline

\end{tabular}
\end{center}
\vspace*{-8pt} \caption{Standard null-move pruning.}
\label{fig:std-algo}
\end{figure}

There are positions in chess where any move will deteriorate the
position, so that not making a move is the best option. These
positions are called \emph{zugzwang} positions. While zugzwang
positions are rare in the middle game, they are not an exception
in endgames, especially endgames in which one or both sides are
left with King and Pawns. Null-move pruning will fail badly in
zugzwang positions since the basic assumption behind the method
does not hold. In fact, the null-move search's value is an upper
bound in such cases. As a result, null-move pruning is avoided in
such endgame positions.

\newpage
As previously noted, the major benefit of null-move pruning
stems from the depth reduction in the null-move searches. However,
these reduced-depth searches are liable to tactical weaknesses due
to the \emph{horizon effect} \citebay{berliner74}. A horizon
effect results whenever the reduced-depth search misses a tactical
threat. Such a threat would not have been missed, had we conducted
a search without any depth reduction. The greater the depth
reduction $R$, the greater the tactical risk due to the horizon
effect. So, the saving resulting from null-move pruning depends on
the depth reduction factor, since a shallower search (i.e., a
greater $R$) will result in faster null-move searches and an
overall smaller search tree.

In the early days of null-move pruning, most programs used $R =
1$, which ensures the least tactical risk, but offers the least
saving in comparison with other $R$ values. Other reduction
factors that were experimented with were $R = 2$ and $R = 3$.
Research conducted over the years, most extensively by Heinz
\citeby{heinz99}, showed that overall, $R = 2$ performs better
than the too conservative $R = 1$ and the too aggressive $R =3$.
Today, almost all null-move pruning programs, use at least $R =
2$~\citebay{feist99}. However, using $R = 3$ is tempting,
considering the reduced search effort resulting from shallower
null-move searches. (This will be demonstrated in Section 4.)
\citeaby{donninger93} was the first to suggest an adaptive rather
than a fixed value for $R$. Experiments conducted by Heinz
\citeby{heinz99}, in his article on adaptive null-move pruning,
suggest that using $R = 3$ in upper parts of the search tree and
$R = 2$ in its lower parts can save 10 to 30 percent of the search
effort in comparison with a fixed $R = 2$, while maintaining
overall tactical strength.

In the next section we present a new null-move pruning method which
allows the use of $R = 3$ in all parts of the search tree, while
alleviating to a significant extent the main disadvantage of standard
null-move pruning.

\section{Verified Null-Move Pruning}

Cutoffs based on a shallow null-move search can be too risky at
some points, especially in zugzwang positions. \citeaby{goetsch90}
hinted at continuing the search with reduced depth, in case the
null-move search indicates a fail-high, in order to substantiate
that the value returned from the null-move search is indeed a
lower bound on the position. \citeaby{plenkner95} showed that this
idea can help prevent errors due to zugzwangs. However, verifying
the search in the middle game seems wasteful, as it appears to
undermine the basic benefit of null-move pruning, namely that a
cutoff is determined by a shallow null-move search.

\begin{figure}[h!]
\begin{center}

\begin{tabular}{|l|}
\hline
\\
\includegraphics{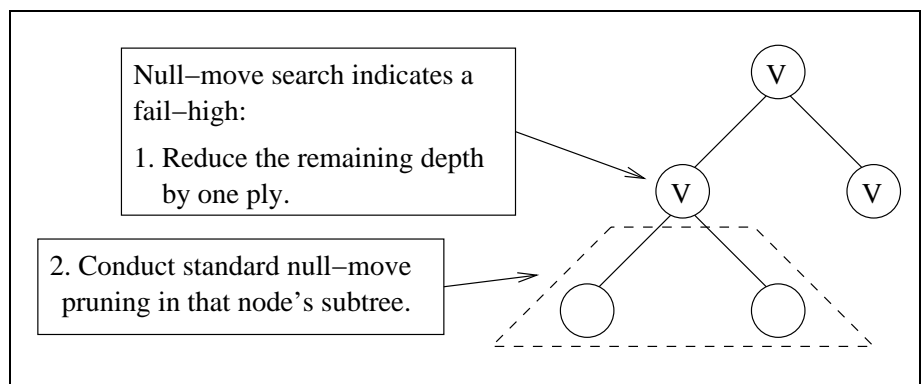}
\\ \\
\hline
\end{tabular}
\end{center}
\vspace{-8pt} \caption{Illustration of verified null-move
pruning.} \label{fig:vrfd-exmpl}

\end{figure}

In addition to helping in detecting zugzwangs, the idea of not
immediately pruning the search tree (based on the value returned
from the shallow null-move search) can also help to reduce the
tactical weaknesses caused by the horizon effect, since by
continuing the search we may be able to detect threats which the
shallow null-move search has failed to detect. Based on these
ideas, we developed our own reformulation, which we call
\emph{verified null-move pruning}. At each node, we conduct a
null-move search with a depth reduction of $R = 3$. If the
returned value from that null-move search indicates a fail-high
(i.e., $value \ge \beta$), we then reduce the depth by one ply and
continue the search in order to verify the cutoff. However, for
that node's subtree, we use standard null-move pruning (cutoff
takes place upon fail-highs). See Figure~\ref{fig:vrfd-exmpl}, for
an illustration.

The basic idea behind verified null-move pruning is that null-move
search with $R = 3$ constructs a considerably smaller search tree.
However, because of its tactical deficiencies, a cutoff based on it is
too risky. So upon a fail-high, we reduce the depth and continue the
search, using standard null-move pruning (with $R = 3$) in that node's
subtree. The search at a node is thus cut off (based on its null-move
search) only if there has been another null-move search fail-high
indication in one of the node's ancestors (see Figure 2). As the
experimental results in the next section show, verified null-move
pruning constructs a search tree which is close in size to that of
standard null-move pruning with $R = 3$, and whose tactical strength
is greater on average than that of standard null-move pruning with
$R = 2$. This is a smaller search tree with greater tactical strength,
in comparison with standard null-move pruning with $R = 2$, which is
commonly used nowadays.

Since upon a fail-high indication we do not cut off the search at once,
we have the ability to check whether the returned value is indeed a lower
bound on the position. If the null-move search indicates a cutoff, but the
search shows that the best value is smaller than $\beta$, this implies
that the position is a zugzwang, as the value from the null move is
greater than or equal to the value from the best move. In such cases,
we restore the original depth (which was reduced by one ply after the
fail-high indication), and conduct a re-search to obtain the correct value.

Implementation of verified null-move search is a matter of adding a few
lines of code to standard null-move search, as shown in
Figure~\ref{fig:vrfd-algo}. Regarding the pseudo-code presented, when the
search starts at the root level, the flag {\tt verify} is initialized to
{\tt true}. When the null-move search indicates a fail-high, the
remaining depth is reduced by one ply, and {\tt verify} is given the
value {\tt false}, which will be passed to the children of the current node,
indicating that standard null-move pruning will be conducted with respect
to the children.  Upon a fail-high indication due to the standard null-move
search of these children's subtrees, cutoff takes place immediately.

\begin{figure}[htbp]
\begin{center}
\begin{tabular}{|l|}
\hline
{\tt \#define R \ 3} {\sl /* the depth reduction factor */}\\
{\sl /* at the root level, verify = true */}\\
{\tt int search (alpha, beta, depth, verify)} \{\\
\tab {\tt if (depth $<$= 0)}\\
\tab \tab {\tt return evaluate();} {\sl /* in practice, quiescence()
is called here */} \\
\tab {\sl /* if verify = true, and depth = 1, null-move search is not
conducted, since}\\
\tab {\sl \ * verification will not be possible */}\\
\tab {\tt if ($!$in\_check() \&\& null\_ok() \&\& ($!$verify || depth
$>$ 1))} \{\\
\tab \tab {\tt make\_null\_move();}\\
\tab \tab {\sl /* null-move search with minimal window around beta */}\\
\tab \tab {\tt value = -search(-beta, -beta + 1, depth - R - 1,}\\
\tab \tab {\tt \tab \tab \hspace*{0.8cm} verify);}\\
\tab \tab {\tt if (value $>$= beta)} \{ {\sl /* fail-high */}\\
\tab \tab \tab {\tt if (verify)} \{\\
\tab \tab \tab \tab {\tt depth--;} {\sl /* reduce the depth by one ply
*/}\\
\tab \tab \tab \tab {\sl /* turn verification off for the sub-tree */}\\
\tab \tab \tab \tab {\tt verify = false;}\\
\tab \tab \tab \tab {\sl /* mark a fail-high flag, to detect zugzwangs
later*/}\\
\tab \tab \tab \tab {\tt fail\_high = true;}\\
\tab \tab \tab \}\\
\tab \tab \tab {\tt else} {\sl /* cutoff in a sub-tree with fail-high
report */}\\
\tab \tab \tab \tab {\tt return value;}\\
\tab \tab \}\\
\tab \}\\
{\tt re\_search:} {\sl /* if a zugzwang is detected, return here for
re-search */}\\
\tab {\sl /* do regular NegaScout/PVS search */}\\
\tab {\sl /* search() is called with current value of ``verify'' */}\\
\tab \ldots \\

\tab {\sl /* if there is a fail-high report, but no cutoff was found, the
position}\\
\tab {\sl \ * is a zugzwang and has to be re-searched with the original
depth */}\\
\tab {\tt if(fail\_high \&\& best $<$ beta)} \{\\
\tab \tab {\tt depth++;}\\
\tab \tab {\tt fail\_high = false;}\\
\tab \tab {\tt verify = true;}\\
\tab \tab {\tt goto re\_search;}\\
\tab \}\\
\}\\
\hline

\end{tabular}
\end{center}
\vspace*{-8pt} \caption{Verified null-move pruning.}
\label{fig:vrfd-algo}
\end{figure}

\section{Experimental Results \label{sec:experiments}}

In this section we examine the performance of verified null-move
pruning, focusing on its tactical strength and smaller search-tree
size in comparison with standard null-move pruning. We conducted
our experiments using the
\textsc{Genesis}\footnote{http://www.omiddavid.com/genesis}
engine. \textsc{Genesis} is designed especially for research,
emphasizing accurate implementation of algorithms and detailed
statistics. For our experiments we used the \textsc{NegaScout}/PVS
\citebay{campbell83,reinefeld83} search algorithm, with history
heuristic (Schaeffer, 1983, 1989) \nocite{schaeffer83,schaeffer89}
and transposition table \citebay{slate77,nelson85}. To demonstrate
the tactical strength differences between the different methods
even better, we used one-ply check extensions on leaf nodes; the
quiescence search consisted only of captures/recaptures. In all
test suites used, we discarded positions in which at least one
side had no more than King and Pawns. This was done to avoid
dealing with zugzwang positions, for which verified null-move
pruning obviously fares much better tactically, as explained
before.

In order to obtain an estimate of the search tree, we searched 138
test positions from \emph{Test Your Tactical Ability} by Yakov
Neishtadt (see the Appendix) to depths of 9 and 10 plies, using
standard $R = 1$, $R = 2$, $R = 3$, and verified $R = 3$.
Table~\ref{tab:nodes-neishtadt} gives the total node count for
each method and the size of the tree in comparison with verified
$R = 3$. Table~\ref{tab:solved-neishtadt} gives the number of
positions that each method solved correctly (i.e., found the
correct variation for). Later we will further examine the tactical
strength, using additional test suites.

\begin{table}[h!]

\begin{center}
\begin{tabular}{|c||r|r|r||r|}
\hline
Depth & Std $R = 1$ & Std $R = 2$ & Std $R = 3$ & Vrfd $R = 3$\\
\hline \hline
9 & 1,652,668,804 & 603,549,661 & 267,208,422 & 449,744,588\\
& (+267.46\%) & (+34.19\%) & (-40.58\%) & -\\
\hline
10 & 11,040,766,367 & 1,892,829,685 & 862,153,828 & 1,449,589,289\\
& (+661.64\%) & (+30.57\%) & (-40.52\%) & -\\
\hline

\end{tabular}
\end{center}
\vspace*{-8pt} \caption{Total node count of standard $R = 1, 2, 3$
and verified $R = 3$ at depths 9 and 10, for 138 Neishtadt test
positions.} \label{tab:nodes-neishtadt}

\end{table}

\begin{table}

\begin{center}
\begin{tabular}{|c||c|c|c||c|}
\hline
Depth & Std $R = 1$ & Std $R = 2$ & Std $R = 3$ & Vrfd $R = 3$\\
\hline \hline
9 & 64 & 62 & 53 & 60\\
\hline
10 & 71 & 66 & 65 & 71\\
\hline
\end{tabular}
\end{center}
\vspace*{-8pt} \caption{Number of solved positions using standard
$R = 1, 2, 3$ and verified $R = 3$ at depths 9 and 10, for 138
Neishtadt test positions.} \label{tab:solved-neishtadt}

\end{table}

\begin{figure}
\begin{center}

\includegraphics{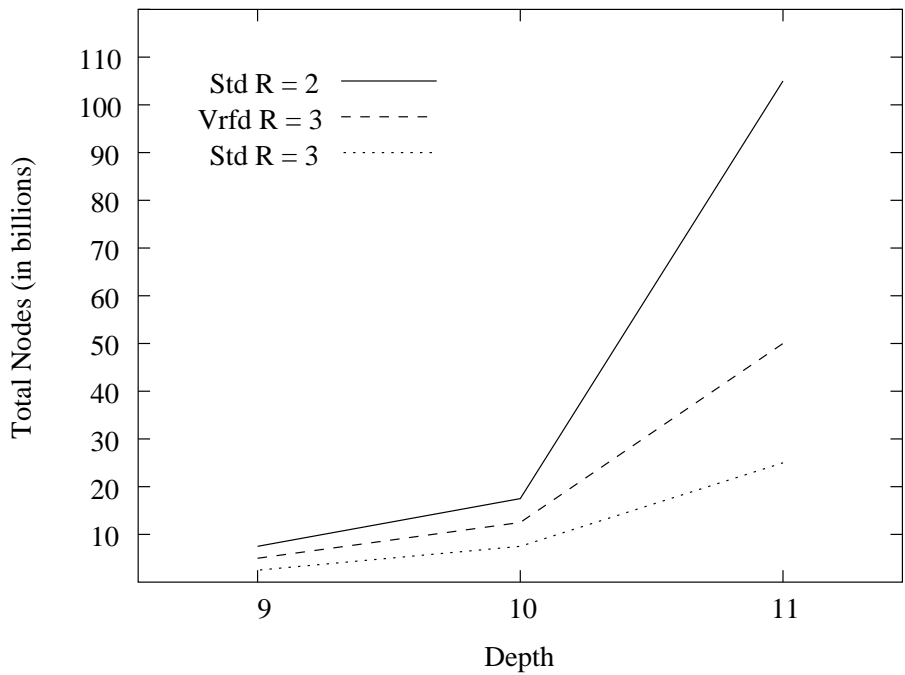}
\end{center}
\vspace*{-8pt} \caption{Tree sizes of standard $R = 2$, $R = 3$,
and verified $R = 3$ at depths 9, 10, and 11, for 869 ECM test
positions.} \label{fig:vrfd-graph}

\end{figure}

The results in Tables \ref{tab:nodes-neishtadt} and
\ref{tab:solved-neishtadt} reveal that the size of the tree
constructed by verified null-move pruning is between those of
standard $R = 2$ and $R = 3$, and that its tactical strength is
greater on average than that of standard $R = 2$. These results
also show that the use of $R = 1$ is impractical due to its large
tree size in comparison with other depth-reduction values.
Focusing on the practical alternatives (i.e., standard $R = 2$ and
$R = 3$, and verified $R = 3$), we would like to examine the
behavior of verified $R = 3$ and find out whether its tree size
remains between the tree sizes associated with $R = 2$ and $R =
3$, or whether it approaches the size of one of these trees. We
therefore conducted a search to a depth of 11 plies, using 869
positions from the \textit{Encyclopedia of Chess Middlegames}
(ECM)\footnote{Because of the large number of errors in ECM's
suggested best moves, we did not check here for number of solved
positions.}. Table~\ref{tab:nodes-ecm} provides the total node
counts at depths 9, 10, and 11, using standard $R = 2$, $R = 3$,
and verified $R = 3$. See also Figure~\ref{fig:vrfd-graph}.


\begin{table}

\begin{center}
\begin{tabular}{|c||r|r||r|}
\hline
Depth & Std $R = 2$ & Std $R = 3$ & Vrfd $R = 3$\\
\hline \hline
9 & 5,374,275,763 & 2,483,951,601 & 4,848,596,820\\
& (+10.84\%) & (-48.76\%) & -\\
\hline
10 & 16,952,333,579 & 7,920,812,800 & 14,439,185,304\\
& (+17.40\%) & (-45.14\%) & -\\
\hline
11 & 105,488,197,524 & 24,644,668,194 & 51,080,338,048\\
& (+106.51\%) &(-51.75\%) & -\\
\hline
\end{tabular}
\end{center}
\vspace*{-8pt} \caption{Total node count of standard $R = 2$, $R =
3$, and verified $R = 3$ at depths 9, 10, and 11, for 869 ECM test
positions.} \label{tab:nodes-ecm}

\end{table}

As Figure~\ref{fig:vrfd-graph} clearly indicates, for depth 11 the
size of the tree constructed by verified null-move pruning with $R
= 3$ is closer to standard null-move pruning with $R = 3$. This
implies that the saving from verified null-move pruning will be
greater as we search more deeply. This can be explained by the
fact that the saving from the use of $R = 3$ in the shallow
null-move search far exceeds the verification cost of verified
null-move pruning.

Having studied the effect of verified null-move pruning on the
search tree size, we now take a closer look at the resulting
tactical strength in comparison with standard null-move pruning
with different depth reductions. For this purpose we used 999
positions from the \textit{Winning Chess Sacrifices} (WCS) test
suite, and 434 positions of ``mate in 4'' and 353 positions of
``mate in 5'' from the test suites of the \textit{Chess Analysis
Project} (CAP); see the Appendix. The WCS positions were searched
to depths of 8, 9, and 10 plies, using standard $R = 2$, $R = 3$,
and verified $R = 3$. Table~\ref{tab:nodes-wcs} provides the total
node counts, and Table~\ref{tab:wcs} gives the number of correctly
solved positions for the WCS test suite. For each position of
``mate in 4'' we conducted a search to a depth of 8 plies, and for
each ``mate in 5'' position a search to a depth of 10 plies. The
search was conducted using standard $R = 1$, $R = 2$, $R = 3$, and
verified $R = 3$. Table~\ref{tab:mates} provides the number of
positions that each method solved (i.e., found the checkmating
sequence).


\begin{table}

\begin{center}
\begin{tabular}{|c||r|r||r|}
\hline
Depth & Std $R = 2$ & Std $R = 3$ & Vrfd $R = 3$\\
\hline \hline
8 & 783,461,647 & 533,282,695 & 906,225,552\\
& (-13.55\%) & (-41.15\%) & -\\
\hline
9 & 3,742,064,688 & 1,316,719,980 & 2,539,057,043\\
& (+47.38\%) & (-48.14\%) & -\\
\hline
10 & 11,578,143,939 & 4,871,295,877 & 7,889,544,754\\
& (+46.75\%) &(-38.26\%) & -\\
\hline
\end{tabular}
\end{center}
\vspace*{-8pt} \caption{Total node count of standard $R = 2$, $R =
3$ and verified $R = 3$ at depths 8, 9, and 10, for 999 WCS test
positions.} \label{tab:nodes-wcs}

\end{table}

\begin{table}

\begin{center}
\begin{tabular}{|c||c|c||c|}
\hline
Depth & Std $R = 2$ & Std $R = 3$ & Vrfd $R = 3$\\
\hline \hline
8 & 762 & 760 & 782\\
\hline
9 & 838 & 812 & 838\\
\hline
10 & 850 & 849 & 866\\
\hline
\end{tabular}
\end{center}
\vspace*{-8pt} \caption{Number of solved positions using $R = 2$,
$R = 3$ and verified $R = 3$ at depths 8, 9, and 10 for 999 WCS
test positions.} \label{tab:wcs}

\end{table}


\begin{table}[h!]

\begin{center}
\begin{tabular}{|c||c|c|c||c|}
\hline
Test Suite & Std $R = 1$ & Std $R = 2$ & Std $R = 3$ & Vrfd $R = 3$\\
\hline \hline
``Mate in 4'' & 433 & 385 & 379 & 431\\
Depth 8 plies & & & &\\
\hline
``Mate in 5'' & 347 & 292 & 286 & 340\\
Depth 10 plies & & & &\\
\hline
\end{tabular}
\end{center}
\vspace*{-8pt} \caption{Numbers of solved positions using standard
$R = 1, 2, 3$ and verified $R = 3$ for 434 ``mate in 4'' and 353
``mate in 5'' test suites.} \label{tab:mates}

\end{table}


The results in Tables \ref{tab:wcs} and \ref{tab:mates} indicate that
verified null-move pruning solved far more positions than standard
null-move pruning with depth reductions of $R = 2$ and $R = 3$.
This demonstrates that not only does verified null-move pruning result
in a reduced search effort (the constructed search tree is closer in size
to that of standard $R = 3$), but its tactical strength is greater than
that of standard $R = 2$, which is the common depth reduction value.

Finally, to study the overall advantage of verified null-move
pruning over standard null-move pruning in practice, we conducted
100 self-play games, using two versions of the \textsc{Genesis}
engine, one with verified $R = 3$ and the other with standard $R =
2$. The time control was set to 60 minutes per game. The version
using verified $R = 3$ scored 68.5 out of 100 (see the Appendix),
which demonstrates the superiority of verified null-move pruning
over the standard version.

\section{Conclusion \label{sec:conclusion}}

In this article we introduced a new null-move pruning method which
outperforms standard null-move pruning techniques, in terms of
reducing the search tree size as well as gaining greater tactical
strength. The idea of not cutting off the search as soon as the
shallow null-move search indicates a fail-high allows verification
of the cutoff, which results in greater tactical accuracy and
prevents errors due to zugzwangs. We showed empirically that
verified null-move pruning with a depth reduction of $R = 3$
constructs a search tree which is closer in size to that of the
tree constructed by standard $R = 3$, and that the saving from the
reduced search effort in comparison with standard $R = 2$ becomes
greater as we search more deeply. We also showed that on average,
the tactical strength of verified null-move pruning is greater
than that of standard null-move pruning with $R = 2$. Moreover,
verified null-move pruning can be implemented within any standard
null-move pruning framework by merely adding a few lines of code.

We considered a number of variants of standard null-move pruning.
The first variant was not to cut off at all upon fail-high
reports, but rather reduce the depth by 2 plies. We obtained good
results with this idea, but its tactical strength was sometimes
smaller than that of standard $R = 2$. We concluded that in order
to improve the results, the depth should not be reduced by more
than one ply at a time upon fail-high reports. An additional
variant was not to cut off at any node, not even in the subtree of
a node with a fail-high report, but merely to reduce the depth by
one ply upon a fail-high report. Unfortunately, the size of the
resulting search tree exceeded the size of the tree constructed by
standard $R = 2$. Still, another variant was to reduce the depth
by one ply upon fail-high reports, and to reduce the depth by two
plies upon fail-high reports in that node's subtree, rather than
cutting off.

Our empirical studies showed that cutting off the search at the
subtree of a fail-high reported node does not decrease tactical
strength. Indeed, this is the verified null-move pruning version
that we studied in this article. In contrast to the standard
approach which advocates the use of immediate cutoff, the novel
approach taken here uses depth reduction, and delays cutting off
the search until further verification. This yields greater
tactical strength and a smaller search tree.

\section{Acknowledgements}

We would like to thank Shay Bushinsky for his interest in our research,
and for promoting the discipline of Computer Chess in our department.
We would also like to thank Dann Corbit for providing the CAP test
positions for our empirical studies, and Azriel Rosenfeld for his
editorial comments. Finally, we are indebted to Jonathan Schaeffer and
Christian Donninger for their enlightening remarks and suggestions.

\section{Appendix}
\section*{Experimental Setup}
Our experimental setup consisted of the following resources:

\begin{itemize}

\item 138 positions (Diagrams 241 to 378) from: Yakov Neishtadt (1993).
\emph{Test Your Tactical Ability}, pp. 110--135. Batsford, ISBN
0-7134-4013-9.

\item 869 positions from \emph{Encyclopedia of Chess Middlegames}, and 999
positions from \emph{Winning Chess Sacrifices}, as available on
the Internet.

\item 434 ``Mate in 4'' and 353 ``Mate in 5'' positions from \emph{Chess
Analysis Project}, available at ftp://cap.connx.com/

\item \textsc{Genesis} chess engine, with $2^{22}$ transposition table
entries (64MB), running on a 733 MHz Pentium III with 256MB RAM, with
the Windows 98 operating system.

\end{itemize}

The webpage http://www.omiddavid.com/pubs.html
contains additional information about the test suites, move lists
of self-play games, and detailed experimental results.

\end{document}